# Inferring Strategies for Sentence Ordering in Multidocument News Summarization

**Regina Barzilay**                                     REGINA@CS.COLUMBIA.EDU
**Noemie Elhadad**                                      NOEMIE@CS.COLUMBIA.EDU
**Kathleen R. McKeown**                                 KATHY@CS.COLUMBIA.EDU
*Columbia University, Computer Science Department*
*1214 Amsterdam Ave*
*New York, 10027, NY, USA*

## Abstract

The problem of organizing information for multidocument summarization so that the generated summary is coherent has received relatively little attention. While sentence ordering for single document summarization can be determined from the ordering of sentences in the input article, this is not the case for multidocument summarization where summary sentences may be drawn from different input articles. In this paper, we propose a methodology for studying the properties of ordering information in the news genre and describe experiments done on a corpus of multiple acceptable orderings we developed for the task. Based on these experiments, we implemented a strategy for ordering information that combines constraints from chronological order of events and topical relatedness. Evaluation of our augmented algorithm shows a significant improvement of the ordering over two baseline strategies.

## 1. Introduction

Multidocument summarization poses a number of new challenges over single document summarization. Researchers have already investigated issues such as identifying repetitions or contradictions across input documents and determining which information is salient enough to include in the summary (Barzilay, McKeown, & Elhadad, 1999; Carbonell & Goldstein, 1998; Elhadad & McKeown, 2001; Mani & Bloedorn, 1997; McKeown, Klavans, Hatzivassiloglou, Barzilay, & Eskin, 1999; Radev & McKeown, 1998; White, Korelsky, Cardie, Ng, Pierce, & Wagstaff, 2001). One issue that has received little attention is how to organize the selected information so that the output summary is coherent. Once all the relevant pieces of information have been selected across the input documents, the summarizer has to decide in which order to present them so that the whole text makes sense. In single document summarization, one possible ordering of the extracted information is provided by the input document itself. However, Jing (1998) observed that, in single document summaries written by professional summarizers, extracted sentences do not always retain their precedence orders in the summary. Moreover, in the case of multiple input documents, this does not provide a useful solution: information may be drawn from different documents and therefore, no single document can provide an ordering. Furthermore, the order between two pieces of information can change significantly from one document to another.

In this paper, we provide a corpus based methodology for studying ordering. Our goal was to develop a good ordering strategy in the context of multidocument summarization targeted for the news genre. The first question we addressed is the importance of ordering. We conducted experiments which show that ordering significantly affects the reader's comprehension of a text. Our experiments also show that although there is no single ideal ordering of information, ordering is not an unconstrained problem; the number of good orderings for a given text is limited. The second question addressed was the analysis and use of data to infer a strategy for ordering. Existing corpus based methods, such as supervised learning, are not easily applicable to our problem in part because of lack of training data. Given that there are multiple possible orderings, a corpus providing one ordering for each set of information does not allow us to differentiate between sentences which must be together and sentences which happen to be together. This led us to develop a corpus of data sets, each of which contains multiple acceptable orderings of a single text. Such a corpus is expensive to construct and therefore, does not provide enough data for pure statistical approaches. Instead, we used a hybrid corpus analysis strategy that first automatically identifies commonalities across orderings. Manual analysis of the resulting clusters led to the identification of constraints on ordering. Finally, we evaluated plausible ordering strategies by asking humans to judge the results.

Our set of experiments together suggests an ordering algorithm that integrates constraints from an approximation of the temporal sequence of the underlying events and relatedness between content elements. Our evaluation of plausible strategies measured the usefulness of a Chronological Ordering algorithm used in previous summarization systems (McKeown et al., 1999; Lin & Hovy, 2001) as well as an alternative, original strategy, Majority Ordering. Our evaluation showed that the two ordering algorithms alone do not yield satisfactory results. The first, Majority Ordering, is critically linked to the level of similarity of information ordering across the input texts. When input texts have different orderings, however, the algorithm produces unpredictable and unacceptable results. The second, Chronological Ordering produces good results when the information is event-based, and therefore, is temporally sequenced. When texts do not refer to events, but describe states or properties, this algorithm falls short.

Our automatic analysis revealed that topical relatedness is an important constraint; groups of related sentences tend to appear together. Our algorithm combines Chronological Ordering with constraints from topical relatedness. Evaluation shows that the augmented algorithm significantly outperforms either of the simpler methods alone. This strategy can be characterized as bottom-up since final ordering of the text emerges from how the data groups together, whether by related content or by chronological sequence. This contrasts with top-down strategies such as RST (Moore & Paris, 1993; Hovy, 1993), schemas (McKeown, 1985) or plans (Dale, 1992) which impose an external, rhetorically motivated ordering on the data.

In the following sections, we first show that the way information is ordered in a summary can critically affect its overall quality. We then give an overview of our summarization system, MULTIGEN. We next describe the two naive ordering algorithms and evaluate them, followed by a study of multiple orderings produced by humans. This allows us to determine how to improve the Chronological Ordering algorithm using cohesion as an additional constraint. The last section describes the augmented algorithm along with its evaluation.

## 2. Impact of Ordering on the Overall Quality of a Summary

Even though the problem of ordering information for multidocument summarization has received relatively little attention, we hypothesize that good ordering is crucial to produce summaries of quality. The consensus architecture of the state of the art summarizers consists of a content selection module in which the salient information is extracted and a regeneration module in which the information is reformulated into a fluent text. Ideally, the regeneration component contains devices that perform surface repairs on the text by doing anaphora resolution, introducing cohesion markers or choosing the appropriate lexical paraphrases. Our claim in this paper is that the multidocument summarization architecture needs an explicit ordering component. If two pieces of information extracted by the content selection phase end up together but should not, in fact, be next one to another, surface devices will not repair the impaired flow of information in the summary. An ordering strategy would help avoid this situation.

It is clear that ordering cannot improve the output of earlier stages of a summarizer, among them content selection[1]; however, finding an acceptable ordering can enhance user comprehension of the summary and, therefore, its overall quality. Of course, surface devices are still needed to smooth the output summary, but this is out of the scope of this paper (but see (Schiffman, Nenkova, & McKeown, 2002)). In this section we show that the quality of ordering has a direct effect on user comprehension of the summary. To verify our hypothesis, we performed an experiment, measuring the impact of ordering on the user's comprehension of summaries.

We selected ten summaries produced by the Columbia Summarization system (McKeown, Barzilay, Evans, Hatzivassiloglou, Kan, Schiffman, & Teufel, 2001). It is composed of a router and two underlying summarizers — MultiGen and DEMS (Difference Engine for Multidocument Summarization). Depending on the type of input articles to be summarized, the router selects the appropriate summarizer. We evaluated this system through the Document Understanding Conference 2001 (DUC)[2] evaluation, where summaries produced by several systems were graded by human judges according to different criteria, among them how well the information contained in the summary is ordered. To actually identify a possible impact of ordering on comprehension, we selected only summaries where humans judged the ordering as poor.[3] For each summary, we manually reordered the sentences generated by the summarizer, using the input articles as a reference. When doing so, we did not change the content — all the sentences in the reordered summaries were the same ones as in the originally produced summaries. This process yields ten additional reordered summaries and thus, overall our collection contains twenty summaries.

Two subjects other than the authors participated in this experiment. Each summary was read by one participant without having access to the input articles. We distributed the summaries among the judges so that none of them read both an original summary and its reordering. They were asked to grade how well the summary could be understood, using the ratings "Incomprehensible," "Somewhat comprehensible" or "Comprehensible".

---

1. No information is added or deleted once the content selection is performed.

2. `http://www-nlpir.nist.gov/projects/duc/`

3. The selected summaries were produced by the DEMS system. We didn't select any summary produced by MultiGen because it implemented our ordering algorithm at the time. DEMS on the other hand, had no specific ordering strategy implemented and thus provided us with the appropriate type of data.

The results are shown in Figure 1[4]. Seven original summaries were considered incomprehensible by their judge, two were somewhat comprehensible, and only one original summary was fully comprehensible. The reordered summaries obtained better grades overall — five summaries were fully comprehensible, two were somewhat comprehensible, while three remained incomprehensible. To assess the statistical significance of our results, we applied the Fisher exact test to our data set, conflating "Incomprehensible" and "Somewhat comprehensible" summaries into one category to obtain a 2x2 table. This test is adapted to our case because of the reduced size of our data set. We obtained a p-value of 0.07 (Siegal & Castellan, 1988), which means that if reordering is not, in general, helpful, there is only a 7% chance that doing reordering anyway would produce a result this different in quality from the original ordering. This experiment indicates that a good ordering can improve the overall comprehensibility of a summary.

| Summary set | Original | Reordered |
| --- | --- | --- |
| d13 | Incomprehensible | Incomprehensible |
| d19 | Somewhat comprehensible | Comprehensible |
| d24 | Incomprehensible | Comprehensible |
| d31 | Somewhat comprehensible | Comprehensible |
| d32 | Incomprehensible | Somewhat comprehensible |
| d39 | Incomprehensible | Incomprehensible |
| d45 | Incomprehensible | Incomprehensible |
| d50 | Incomprehensible | Comprehensible |
| d54 | Incomprehensible | Somewhat comprehensible |
| d56 | Comprehensible | Comprehensible |

Figure 1: Impact of ordering on the user comprehension of summaries.

In the case of some low-scoring summaries, it is clear that poor ordering is the likely culprit. For instance, readers can easily identify that grouping the two following sentences is an unsuitable choice and could be misleading. *"Miss Taylor's health problems started with a fall from a horse when she was 13 and filming the movie National Velvet. The recovery of Elizabeth Taylor, near death two weeks ago with viral pneumonia, was complicated by a yeast infection, her doctors said Friday."* But in other cases, when information in a summary is poorly ordered and readers cannot make sense of the text, we observed through interviews with the readers that they tend to blame it on content selection rather than on ordering, even if the content is not the issue. Thus, the issue of ordering is not isolated; it can affect the overall quality of a summary.

## 3. MULTIGEN Overview

Our framework is the MULTIGEN system (McKeown et al., 1999), a multidocument summarizer which has been trained and tested on news articles. MULTIGEN is part of the Columbia Summarization System. It operates on a set of news articles describing the same event, creating a summary which synthesizes common information across documents. The system runs daily over real data within Newsblaster[5], a tool which collects news articles from multiple sources, organizes them into topical clusters and provides a summary for each of the clusters.

---

4. The set names are the ones used in the DUC evaluation.

In the case of multidocument summarization of articles about the same event, source articles can contain both repetitions and contradictions. Extracting all the similar sentences would produce a verbose and repetitive summary, while extracting only some of the similar sentences would produce a summary biased towards some sources. MULTIGEN uses a comparison of extracted similar sentences to select the appropriate phrases to include in the summary and reformulates them as new text.

MULTIGEN consists of an analysis and a generation component. The analysis component (Hatzivassiloglou, Klavans, & Eskin, 1999) identifies units of text which convey similar information across the input documents using statistical techniques and shallow text analysis. Once similar text units are identified, we cluster them into *themes*. Themes are sets of sentences from different documents that contain repeated information and do not necessarily contain sentences from all the documents (see two examples of themes in Figure 2). For each theme, the generation component (Barzilay et al., 1999) identifies phrases which are in the intersection of the theme sentences and selects them as part of the summary. The intersection sentences are then ordered to produce a coherent text. At the end, for each theme there will be a single corresponding generated output sentence in the summary. In the following section, we describe different strategies for ordering the output sentences to obtain a quality summary.

| Theme 1 |
| --- |
| Mr. Salvi, 24, apparently killed himself in his prison cell last November. |
| The state wouldn't execute him for killing two abortion clinic workers in 1994, so John C. Salvi III took his own life. |
| John C. Salvi III, who was convicted of killing two people in a shooting spree on two abortion clinics in 1994, killed himself in prison. |

| Theme 2 |
| --- |
| His attorneys said he attempted suicide twice before in prison. |
| His lawyers said that he twice had tried to commit suicide in jail, a charge authorities have denied. |

Figure 2: Two themes with their corresponding sentences. Theme 2 contains sentences from only two articles, while Theme 1 contains sentences from three input articles.

## 4. Naive Ordering Algorithms Are Not Sufficient

When producing a summary, any multidocument summarization system has to choose in which order to present the output sentences. In this section, we describe two algorithms for ordering sentences suitable for multidocument summarization in the news genre. The first algorithm, Majority Ordering (MO), relies only on the original orders of sentences in the input documents. The second one, Chronological Ordering (CO), uses time-related features to order sentences. This strategy was originally implemented in MULTIGEN and followed by other summarization systems (Radev, Jing, & Budzikowska, 2000; Lin & Hovy, 2001). In the MULTIGEN framework, ordering sentences is equivalent to ordering themes, and we describe the algorithms in terms of themes. This makes sense because, ultimately, the summary will be composed of a sequence of sentences, each one constructed from the information in one theme. Our evaluation shows that these methods alone do not provide an adequate strategy for ordering.

---

5. http://www.cs.columbia.edu/nlp/newsblaster

## 4.1 Majority Ordering

### 4.1.1 THE ALGORITHM

In single document summarization, the order of sentences in the output summary is typically determined by their order in the input text. This strategy can be adapted to multidocument summarization. Consider two themes, $Th_1$ and $Th_2$; if sentences from $Th_1$ precede sentences from $Th_2$ in all input texts, then presenting $Th_1$ before $Th_2$ is likely to be an acceptable order. To use the majority ordering algorithm when the order between sentences from $Th_1$ and $Th_2$ varies from one text to another, we must augment the strategy. One way to define the order between $Th_1$ and $Th_2$ is to adopt the order occurring in the majority of the texts where $Th_1$ and $Th_2$ occur. This strategy defines a pairwise order between themes. However, this pairwise relation is not necessarily transitive. For example, given the themes $Th_1$, $Th_2$ and $Th_3$ and the following situation: $Th_1$ precedes $Th_2$ in a text, $Th_2$ precedes $Th_3$ in the same text or in another text, and $Th_3$ precedes $Th_1$ in yet another text; there is a conflict between the orders $(Th_1, Th_2, Th_3)$ and $(Th_3, Th_1)$. Since transitivity is a necessary condition for a relation to be called an order, this relation does not form an order.

We, therefore, have to expand this pairwise relation to provide a total order. In other words, we have to find a linear ordering between themes which maximizes the agreement between the orderings provided by the input texts. For each pair of themes, $Th_i$ and $Th_j$, we keep two counts, $C_{i,j}$ and $C_{j,i}$; $C_{i,j}$ is the number of input texts in which sentences from $Th_i$ occur before sentences from $Th_j$, and $C_{j,i}$ is the same for the opposite order. The weight of a linear order $(Th_{i_1}, \ldots, Th_{i_k})$ is defined as the sum of the counts for every pair $C_{i_l, i_m}$, such that $i_l \leq i_m$ and $l, m \in \{1 \ldots k\}$. Stating this problem in terms of a directed graph where nodes are themes, and a vertex from $Th_i$ to $Th_j$ has the weight $C_{i,j}$, we are looking for a path with maximal weight which traverses each node exactly once (see Figure 3). We call such a graph a precedence graph.

The problem of finding a path with maximal weight has been addressed by Cohen, Schapire, and Singer (1999) in the task of learning orderings. They adopt a two-stage approach. In the first stage, given a training corpus of ordered instances and a set of features describing them, a binary preference function is learned. In the second stage, new instances are ordered so that agreement with the learned preference function is maximized. To do so, Cohen et al. (1999) represent the preference function as a directed, weighted graph. Our precedence graph can be seen as such a graph where the preference function between the nodes $Th_i$ and $Th_j$ is $C_{i,j}$. The orderings from the input articles provide us directly with a preference function and, therefore, we do not need to learn it.

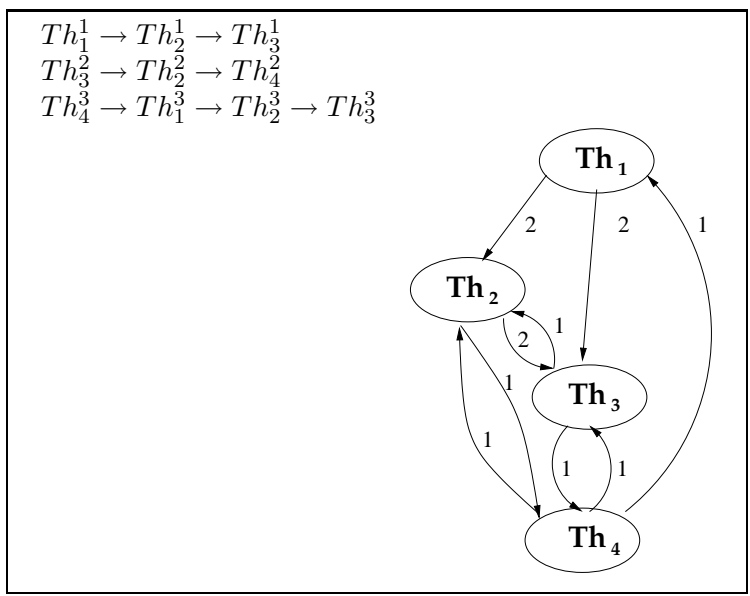

Figure 3: Three input theme orderings and their corresponding precedence graph. $Th_i^j$ is the sentence part of the theme $Th_i$ in the input ordering $j$.

Unfortunately this problem is NP-complete; Cohen et al. (1999) prove it by reducing from CYCLIC-ORDERING (Galil & Megido, 1977). However, using a modified version of topological sort provides us with an approximate solution. For each node, we assign a weight equal to the sum of the weights of its outgoing edges minus the sum of the weights of its incoming edges. We first pick up the node with maximum weight, ordering it ahead of the other nodes, delete it and its outgoing edges from the precedence graph and update properly the weights of the remaining nodes in the graph. We then iterate through the nodes until the graph is empty. Cohen et al. (1999) show that this algorithm produces a tight approximation of the optimal solution. Currently MULTIGEN uses an implementation of this algorithm for its ordering component.

Figures 4 and 5 show examples of produced summaries. One feature of this strategy is that it can produce several orderings with the same weight. This happens when there is a tie between two opposite orderings. In this situation, this strategy does not provide enough constraints to determine one optimal ordering; an ordering is chosen randomly among the orders with maximal weight.

### 4.1.2 EVALUATION

We asked three human judges (not including ourselves) to classify the quality of the order of information in 25 summaries produced using the MO algorithm into three categories—Poor, Fair and Good. We use an operational definition of a Poor summary as a text whose The man accused of firebombing two Manhattan subways in 1994 was convicted Thursday after the jury rejected the notion that the drug Prozac led him to commit the crimes.
He was found guilty of two counts of attempted murder, 14 counts of first-degree assault and two counts of criminal possession of a weapon.
In December 1994, Leary ignited firebombs on two Manhattan subway trains. The second blast injured 50 people – 16 seriously, including Leary.
Leary wanted to extort money from the Transit Authority.
The defense argued that Leary was not responsible for his actions because of "toxic psychosis" caused by the Prozac.

Figure 4: A summary produced using the Majority Ordering algorithm, graded as Good.

Hemingway, 69, died of natural causes in a Miami jail after being arrested for indecent exposure.
A book he wrote about his father, "Papa: A Personal Memoir," was published in 1976.
He was picked up last Wednesday after walking naked in Miami.
"He had a difficult life."
A transvestite who later had a sex-change operation, he suffered bouts of drinking, depression and drifting, according to acquaintances.
"It's not easy to be the son of a great man," Scott Donaldson, told Reuters.
At the time of his death, he lived in the Coconut Grove district where he was well-known to its Bohemian crowd.
He had been due to appear in court later that day on charges of indecent exposure and resisting arrest.
He sometimes went by the name of Gloria and wore women's clothes.
The cause of death was hypertension and cardiovascular disease.
Taken to the Miami-Dade Women's Detention Center, he was found dead in his cell early on Monday, spokeswoman Janelle Hall said.
He was booked into the women's jail because he had a sex-change operation, Hall added.

Figure 5: A summary produced using the Majority Ordering algorithm, graded as Poor.

readability would be significantly improved by reordering its sentences. A Fair summary is a text which makes sense, but reordering of some sentences can yield a better readability. Finally, a summary which cannot be further improved by any sentence reordering is considered a Good summary.

The judges were asked to grade the summaries taking into account only the order in which the information is presented. To help them focus on this aspect of the texts, we resolved dangling references beforehand. Figure 13 shows the grades assigned to the summaries — three summaries were graded as Poor, 14 were graded as Fair, and eight were graded as Good. We are showing here the majority grade that is selected by at least two judges. This was made possible because in our experiments, judges had strong agreement; they never gave three different grades to a summary.

The MO algorithm produces a small number of Good summaries, but most of the summaries were graded as Fair. For instance, the summary graded Good shown in Figure 4 orders the information in a natural way; the text starts with a sentence summary of the event, then the outcome of the trial is given, a reminder of the facts that caused the trial and a possible explanation of the facts. Looking at the Good summaries produced by MO, we found that it performs well when the input articles follow the same order when presenting the information. In other words, the algorithm produces a good ordering if the input articles' orderings have high agreement.

On the other hand, when analyzing Poor summaries, we observed that the input texts have very different orderings. By trying to maximize the agreement of the input texts' orderings, MO produces a new ordering that does not occur in any input text. The ordering is, therefore, not guaranteed to be acceptable. An example of a new produced ordering is given in Figure 5. The summary would be more readable if several sentences were moved around. An example of a better ordering is given in Figure 6. In this summary, the three sentences related to the fact that the subject had a sex-change operation are grouped together, while in the one produced by the majority ordering algorithm, they are scattered throughout the summary.

---

Hemingway, 69, died of natural causes in a Miami jail after being arrested for indecent exposure.
The cause of death was hypertension and cardiovascular disease.
He was picked up last Wednesday after walking naked in Miami.
He had been due to appear in court later that day on charges of indecent exposure and resisting arrest.
Taken to the Miami-Dade Women's Detention Center, he was found dead in his cell early on Monday, spokeswoman Janelle Hall said.
He was booked into the women's jail because he had a sex-change operation, Hall added.
A transvestite who later had a sex-change operation, he suffered bouts of drinking, depression and drifting, according to acquaintances.
He sometimes went by the name of Gloria and wore women's clothes.
"He had a difficult life."
"It's not easy to be the son of a great man," Scott Donaldson, told Reuters.
At the time of his death, he lived in the Coconut Grove district where he was well-known to its Bohemian crowd.
A book he wrote about his father, "Papa: A Personal Memoir," was published in 1976.

---

Figure 6: One possible better ordering for the summary graded as Poor.

This algorithm can be used to order sentences accurately if we are certain that the input texts follow similar organizations. This assumption may hold in limited domains where documents have a fixed organization of the information. However, in our case, the input texts we are processing do not have such regularities. Looking at the daily statistics of Newsblaster which collects clusters of related articles to be synthesized into one summary, we notice that the typical cluster size is seven. But every day there are several clusters which contain more than 20 and up to 70 articles to be summarized into single summaries[6]. With such a big number of input articles, we cannot assume that they will all have similar ordering of the information. MO's performance critically depends on the agreement of orderings in the input texts; we, therefore, need an ordering strategy which can fit any input data. From here on, we will focus only on the Chronological Ordering algorithm and techniques to improve it.

---

6. These giant clusters correspond to the "hot topics" of the day in the news.

## 4.2 Chronological Ordering

### 4.2.1 THE ALGORITHM

Multidocument summarization of news typically deals with articles published on different dates, and articles themselves cover events occurring over a wide range of time. Using chronological order in the summary to describe the main events helps the user understand what has happened. It seems like a natural and appropriate strategy. As mentioned earlier, in our framework, we are ordering themes; using this strategy, we, therefore, need to assign a date to themes. To identify the date an event occurred requires a detailed interpretation of temporal references in articles. While there have been recent developments in disambiguating temporal expressions and event ordering (Wiebe, O'Hara, Ohrstrom-Sandgren, & McKeever, 1998; Mani & Wilson, 2000; Filatova & Hovy, 2001), correlating events with the date on which they occurred is a hard task. In our case, we approximate the theme time by its first publication time; that is, the first time the theme has been reported in our set of input articles (see Figure 7). It is an acceptable approximation for news events; the first publication time of an event usually corresponds to its occurrence in real life. For instance, in a terrorist attack story, the theme conveying the attack itself will have a date previous to the date of the theme describing a trial following the attack.

| Theme 5 | |
|---|---|
| Oct 5, 11:35am | Hours after the crash, U.S. officials said that the tragedy had been caused by an S-200 missile fired by Ukraine during military exercises on the Crimean Peninsula. |
| Oct 6, 6:13am | U.S. officials said immediately after the crash that they had evidence the passenger jet was hit by a Ukrainian missile. |
| <u>Oct 5, 10:20am</u> | But U.S. officials said that the crash had been caused by an S-200 missile fired mistakenly by Ukrainian forces during military exercises on the Crimean Peninsula. |

Figure 7: A theme with its corresponding sentences. The time theme is shown underlined; it is the earliest publication time of the sentences.

Articles released by news agencies are marked with a publication time, consisting of a date and a time with two fields (hour and minutes). Articles from the same news agency are thus guaranteed to have different publication times. This is also quite likely for articles coming from different news agencies. During the development of MULTIGEN, we processed hundreds of articles, and we never encountered two articles with the same publication time. Thus, the publication time serves as a unique identifier over articles. As a result, when two themes have the same publication time, it means that they both are reported for the first time in the same article.

Our Chronological Ordering (CO) algorithm takes as input a set of themes and orders them chronologically whenever possible. Each theme is assigned a date corresponding to its first publication. To do so, we select for each theme the sentence that has the earliest publication time. We call it the time stamp sentence and assign its publication time as the time stamp of the theme. This establishes a partial order over the themes. When two themes have the same date (that is, they are reported for the first time in the same article) we sort them according to their order of presentation in this article. This results in a total order over the input themes. Figures 8 and 9 show examples of summaries produced using CO.

---

One of four people accused along with former Pakistani Prime Minister Nawaz Sharif has agreed to testify against him in a case involving possible hijacking and kidnapping charges, a prosecutor said Wednesday.
Raja Quereshi, the attorney general, said that the former Civil Aviation Authority chairman has already given a statement to police.
Sharif's lawyer dismissed the news when speaking to reporters after Sharif made an appearance before a judicial magistrate to hear witnesses give statements against him. Sharif has said he is innocent.
The allegations stem from an alleged attempt to divert a plane bringing army chief General Pervez Musharraf to Karachi from Sri Lanka on October 12.

---

Figure 8: A summary produced using the Chronological Ordering algorithm graded as Good.

---

Thousands of people have attended a ceremony in Nairobi commemorating the first anniversary of the deadly bombings attacks against U.S. Embassies in Kenya and Tanzania.
Saudi dissident Osama bin Laden, accused of masterminding the attacks, and nine others are still at large.
President Clinton said, "The intended victims of this vicious crime stood for everything that is right about our country and the world".
U.S. federal prosecutors have charged 17 people in the bombings.
Albright said that the mourning continues.
Kenyans are observing a national day of mourning in honor of the 215 people who died there.

---

Figure 9: A summary produced using the Chronological Ordering algorithm graded as Poor.

### 4.2.2 Evaluation

Following the same methodology we used for the MO algorithm evaluation, we asked three human judges (not including ourselves) to grade 25 summaries generated by the system using the CO algorithm applied to the same collection of input texts. The results are shown in Figure 13: ten summaries were graded as Poor, eight were graded as Fair and seven were graded as Good.

Our first suspicion was that our approximation deviates too much from the real chronological order of events and, therefore, lowers the quality of sentence ordering. To verify this hypothesis, we identified sentences that broke the original chronological order and restored the ordering manually. Interestingly, the displaced sentences were mainly background information. The evaluation of the modified summaries shows no visible improvement.

When comparing Good (Figure 8) and Poor (Figure 9) summaries, we notice two phenomena: first, many of the badly placed sentences cannot be ordered based on their temporal occurrence. For instance, in Figure 9, the sentence quoting Clinton is not one event in the sequence of events being described, but rather, a reaction to the main events. A tool assigning time stamps would assign to this sentence the date at which Clinton made his statement. This is also true for the sentence reporting Albright's reaction. Assigning a date to a reaction, or more generally to any sentence conveying background information, and placing it into the chronological stream of the main events does not produce a logical ordering. The ordering of these themes is, therefore, not covered by the CO algorithm. Furthermore, some sentences cannot be assigned any time stamp. For instance, the sentence, *"The vast, sparsely inhabited Xinjiang region, largely desert, has many Chinese military and nuclear installations and civilian mining."* describes a state rather than an event and, therefore, trying to describe it in temporal terms is invalid. Thus the ordering cannot be improved at the temporal level.

The second phenomenon we observed is that Poor summaries typically contain abrupt switches of topics and are generally incoherent. For instance, in Figure 9, quotes from US officials (third and fifth sentences) are split, and sentences about the mourning (first and sixth sentences) appear too far apart in the summary. Grouping them together would increase the readability of the summary. At this point, we need to find additional constraints to improve the ordering.

## 5. Improving the Ordering: Experiments and Analysis

In the previous section, we showed that using naive ordering algorithms does not produce satisfactory orderings. In this section, we investigate through experiments with humans how to identify patterns of orderings that can improve the algorithm.

### 5.1 Collecting a corpus of multiple orderings

Sentences in a text can be ordered in a number of ways, and the text as a whole will still convey the same meaning. But the majority of possible orders are likely to be unacceptable because they break conventions of information presentation. One way to identify these conventions is to find commonalities among different acceptable orderings of the same information. Extracting regularities in several acceptable orderings can help us specify ordering constraints for a given input type. There is no naturally occurring existing collection of summaries for multiple documents that we aware of[7]. But even such a collection would not be sufficient since we want to analyze a collection of multiple summaries over the same set of articles. We created our own collection of multiple orderings produced by different humans. Using this collection, we studied common behaviors and mapped them to strategies for ordering.

Our collection of multiple orderings, along with our test corpus is available at `http://www.cs.columbia.edu/~noemie/ordering/`. We collected ten sets of articles for this collection. Each set consisted of two to three news articles reporting the same event. For each set, we manually selected the intersection sentences, simulating MULTIGEN[8]. On average, each set contained 8.8 intersection sentences. The sentences were cleaned of explicit references (for instance, occurrences of "the President" were resolved to "President Clinton") and connectives, so that participants would not use them as clues for ordering. Ten subjects participated in the experiment, and they each built one ordering per set of intersection sentences. Each subject was asked to order the intersection sentences of a set so that they form a readable text. Overall, we obtained 100 orderings, ten alternative orderings per set. Figure 10 shows the ten alternative orderings collected for one set.

---

7. In a recent attempt, NIST for the DUC conference collected sets of articles to summarize and one summary per set.

8. We performed a manual simulation to ensure that ideal data was provided to the subjects of the experiments.

| | |
|---|---|
| Participant 1 | <u>D B G I</u> H F <u>C J A</u> E |
| Participant 2 | <u>D G B I C F A J E H</u> |
| Participant 3 | <u>D B I G</u> F <u>J A</u> E H C |
| Participant 4 | D <u>C F G</u> I B <u>J A</u> H <u>E</u> |
| Participant 5 | <u>D G B I</u> H F <u>J A</u> C E |
| Participant 6 | <u>D G I B F C</u> E H <u>J A</u> |
| Participant 7 | <u>D B G I F C</u> H E <u>J A</u> |
| Participant 8 | D B <u>C F</u> G I E H <u>A J</u> |
| Participant 9 | <u>D G I B E H</u> F A J C |
| Participant 10 | <u>D B G I C F A J E H</u> |

Figure 10: Multiple orderings for one set in our collection. A, B, ... J stand for sentences. Underlined are automatically identified blocks.

We first observed that a surprisingly large portion of the orderings are different. Out of the ten sets, only two sets had some identical orderings (in one set, two orderings were identical while in the other set, there were two pairs of identical orderings). This variety in the produced orderings can be interpreted as suggesting that not all the orderings were actually valid or that the task was maybe too hard for the subjects to allow them to produce reasonable orderings. In fact, all the subjects were satisfied with the orderings they produced. Furthermore, we manually went through all the 100 orderings, and all appeared to be valid. In other words, there are many acceptable orderings given one set of sentences. This confirms the intuition that we do not need to look for a single ideal total ordering but rather construct an acceptable one.

Looking at these various orderings, one might also conclude that any ordering would do just as well as any other. One piece of evidence against this statement is that, as shown in section 2, some orderings yield incomprehensible texts and thus should be avoided. Furthermore, for a text with $n$ sentences, there are $n!$ possible orderings, but only a small fraction of those are actually valid orderings. One way to validate this claim would be to enumerate all the possible orderings of a single text and evaluate each one of them. This would be doable for very small texts (a text of 5 sentences has 120 possible orderings) but not for texts of a reasonable size. A more feasible way to validate our claim is to get multiple orderings of the same text from a large number of subjects. We asked subjects to order one text of eight sentences. There is a maximum of 40,320 possible orderings for these sentences. While 50 subjects participated, we only obtained 21 unique orderings, showing that the number of acceptable orderings does not grow as fast as the number of participants. We can conclude that only a small fraction of all possible orderings of the information in a text contains orderings that render a readable text.

## 5.2 Analysis

The several alternative orderings produced for a single summary exhibit commonalities. We noticed that, within the multiple orderings of a set, some sentences always appear together. They do not appear in the same order from one ordering to another, but they share an adjacency relation. From now on, we refer to them as blocks. For each set, we identify blocks by automatically clustering sentences across orderings. We use as a distance metric between two sentences, the average number of sentences that separate them over all orderings. In Figure 10, for instance, the distance between sentences D and G is 2. The blocks identified by clustering are: sentences B, D, G and I; sentences A and J; sentences C and F; and sentences E and H.

We observed that all the blocks in the experiment correspond to clusters of topically related sentences. These blocks form units of text dealing with the same subject. In other words, all valid orderings contain blocks of topically related sentences. The notion of grouping topically related sentences is known as cohesion. As defined by Hasan (1984), cohesion is a device for "sticking together" different parts of the text. Studies show that the level of cohesion has a direct impact on reading comprehension (Halliday & Hasan, 1976). Therefore, good orderings are cohesive; this is what makes the summary readable. Conversely, the evaluation of the CO algorithm showed that the summaries that were judged invalid contain abrupt switches of topic. In other words, orderings that are not cohesive are graded poorly. There is a correlation between the quality of the ordering and cohesion. Incorporating cohesion constraint into our ordering strategy by opportunistically grouping sentences together would be beneficial. Cohesion is achieved by surface devices, such as repetition of words and coreferences. We describe next how we include cohesion in the CO algorithm based on these surface features.

## 6. The Augmented Algorithm

Disfluencies arise in the output of the CO algorithm when topics are distributed over the whole text, violating cohesion properties (McCoy & Cheng, 1991). A typical scenario is illustrated in Figure 11. The inputs are texts $T_1, T_2, T_3$ (ordered by publication time). $A_1$, $A_2$ and $A_3$ belong to the same theme, whose intersection sentence is $A$, and similarly for $B$ and $C$. The themes $A$ and $B$ are topically related, but $C$ is not related. Summary $S_1$, based only on chronological clues, contains two topical shifts; from $A$ to $C$ and back from $C$ to $B$. A better summary would be $S_2$, which keeps $A$ and $B$ together.

### 6.1 The Algorithm

Our goal is to remove disfluencies from the summary by grouping together topically related themes. The main technical difficulty in incorporating cohesion in our ordering algorithm is to identify and to group topically related themes across multiple documents. In other words, given two themes, we need to determine if they belong to the same cohesion block. For a single document, topical segmentation (Hearst, 1994) could be used to identify blocks, but this technique is not a possibility for identifying cohesion between sentences across multiple documents. Segmentation algorithms typically exploit the linear structure of an input text; in our case, we want to group together sentences belonging to different texts.

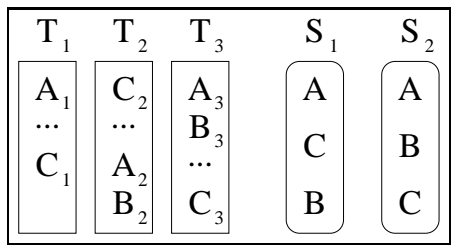

Figure 11: Input texts $T_1 T_2 T_3$ are summarized by the Chronological Ordering ($S_1$) or by the Augmented algorithm ($S_2$).

Our solution consists of the following steps. In a preprocessing stage, we segment each input text (Kan, Klavans, & McKeown, 1998) based on word distribution and coreference analysis, so that given two sentences within the same text, we can determine if they are topically related. Assume the themes $A$ and $B$ exist, where $A$ contains sentences $(A_1 \dots A_n)$, and $B$ contains sentences $(B_1 \dots B_m)$. Recall that a theme is a set of sentences conveying similar information drawn from different input texts. We denote $\#AB$ to be the number of pairs of sentences $(A_i, B_j)$ which appear in the same text, and $\#AB^+$ to be the number of sentence pairs which appear in the same text and are in the same segment.

In the first stage, for each pair of themes $A$ and $B$, we compute the ratio $\#AB^+/\#AB$ to measure the relatedness of two themes. This measure takes into account both positive and negative evidence. If most of the sentences in $A$ and $B$ that appear together in the same texts are also in the same segments, it means that $A$ and $B$ are highly topically related. In this case, the ratio is close to 1. On the other hand, if among the texts containing sentences from $A$ and $B$, only a few pairs are in the same segments, then $A$ and $B$ are not topically related. Accordingly, the ratio is close to 0. $A$ and $B$ are considered related if this ratio is higher than a predetermined threshold. We determined experimentally its value to be 0.6.

This strategy defines pairwise relations between themes. A transitive closure of this relation builds groups of related themes and, as a result, ensures that themes that do not appear together in any article but which are both related to a third theme will still be linked. This creates an even higher degree of relatedness among themes. Because we use a threshold to establish pairwise relations, the transitive closure does not produce elongated chains that could link together unrelated themes. We are now able to identify topically related themes. At the end of the first stage, they are grouped into blocks.

In a second stage, we assign a time stamp to each block of related themes using the earliest time stamp of the themes it contains. We adapt the CO algorithm described in 4.2.1 to work at the level of the blocks. The blocks and the themes correspond to, respectively, themes and sentences in the CO algorithm. By analogy, we can easily show that the adapted algorithm produces a complete order of the blocks. This yields a macro-ordering of the summary. We still need to order the themes inside each block.

In the last stage of the augmented algorithm, for each block, we order the themes it contains by applying the CO algorithm to them. Figure 12 shows an example of a summary produced by the augmented algorithm.

This algorithm ensures that cohesively related themes will not be spread over the text and decreases the number of abrupt switches of topics. Figure 12 shows how the Augmented algorithm improves the sentence order compared with the order in the summary produced by the CO algorithm in Figure 9; sentences quoting US officials are now grouped together, and so are the descriptions of the mourning.

---

Thousands of people have attended a ceremony in Nairobi commemorating the first anniversary of the deadly bombings attacks against U.S. Embassies in Kenya and Tanzania. Kenyans are observing a national day of mourning in honor of the 215 people who died there.

Saudi dissident Osama bin Laden, accused of masterminding the attacks, and nine others are still at large. U.S. federal prosecutors have charged 17 people in the bombings.

President Clinton said, "The intended victims of this vicious crime stood for everything that is right about our country and the world". Albright said that the mourning continues.

---

Figure 12: A summary produced using the Augmented algorithm. Related sentences are grouped into paragraphs.

## 6.2 Evaluation

Following the same methodology used to evaluate the MO and the CO algorithms, we asked the judges to grade 25 summaries produced by the Augmented algorithm. Results are shown in Figure 13.

The manual effort needed to compare and judge system output is extensive considering that each human judge had to read three summaries for each input set as well as skim the input texts to verify that no misleading information was introduced in the summaries. We collected a corpus of 25 sets of articles for evaluation. Overall, there were 75 summaries to be evaluated. The size of our corpus is comparable with the collection used for the DUC evaluation (30 sets of articles). This evaluation shows a significant improvement in the quality of the orderings from the CO algorithm to the Augmented algorithm. To assess the significance of the improvement, we used the Fisher exact test, conflating Poor and Fair summaries into one category (p-value of 0.04). The augmented algorithm also shows an improvement over the MO algorithm (p-value of 0.07).

|  | Poor | Fair | Good |
|---|---|---|---|
| Majority Ordering | 3 | 14 | 8 |
| Chronological Ordering | 10 | 8 | 7 |
| Augmented Ordering | 3 | 8 | 14 |

Figure 13: Evaluation of the the Majority Ordering, the Chronological Ordering and the Augmented Ordering.

## 7. Related Work

Finding an acceptable ordering has not been studied before in domain independent text summarization. In single document summarization, summary sentences are typically arranged in the same order that they were found in the full document, although Jing (1998) reports that human summarizers do sometimes change the original order. In multidocument summarization, the summary consists of fragments of text or sentences that were selected from different texts. Thus, there is no complete ordering of summary sentences that can be found in the original documents.

In domain dependent summarization, it is possible to establish possible orderings *a priori*. A valid ordering is traditionally derived from a manual analysis of a corpus of texts in the domain, and it typically operates over a set of semantic concepts. A semantic representation of the information is usually available as input to the ordering component. For instance, in the specific domain of news on the topic of terrorist attacks, summaries can be constructed by first describing the place of the attack, followed by the number of casualties, who the possible perpetrators are, etc.

Another alternative when ordering information, still in the domain dependent framework, is to use a more data driven approach, which produces a more flexible output. *A priori* defined simple ordering strategies are combined together by looking at a set of features from the input. Elhadad and McKeown (2001) use such techniques to produce patient specific summaries of technical medical articles. Examples of features which influence the ordering are presence of contradiction or repetition, relevance to the patient characteristics, or type of results being reported. A linear combination of these features assigns a weight to each semantic predicate to be included in the output, allowing them to be ordered. In this case, the features are domain dependent and have been identified through corpus analysis and interviews with physicians. In the case of a domain independent system, it would be an entire new challenge to define and compute such a set of features.

Producing a good ordering of information is also a critical task for the generation community, which has extensively investigated the issue (McKeown, 1985; Moore & Paris, 1993; Hovy, 1993; Bouayad-Agha, Power, & Scott, 2000; Mooney, Carberry, & McCoy, 1990). One approach is top-down, using schemas (McKeown, 1985) or plans (Dale, 1992) to determine the organizational structure of the text. This approach postulates a rhetorical structure which can be used to select information from an underlying knowledge base. Because the domain is limited, an encoding can be developed of the kinds of propositional content that match rhetorical elements of the schema or plan, thereby allowing content to be selected and ordered. Rhetorical Structure Theory (RST) allows for more flexibility in ordering content by establishing relations between pairs of propositions. Constraints based on intention (e.g., Moore & Paris, 1993), plan-like conventions (e.g., Hovy, 1993), or stylistic constraints (e.g., Bouayad-Agha et al., 2000) are used as preconditions on the plan operators containing RST relations to determine when a relation is used and how it is ordered with respect to other relations. Another approach (Mooney et al., 1990) is bottom-up and is used to group together stretches of text in a long, generated document by finding propositions that are related by a common focus. Since this approach was developed for a generation system, it finds related propositions by comparisons of proposition arguments at the semantic level.

In our case, we are dealing with a surface representation, so we find alternative methods for grouping text fragments.

A more recent approach by Duboue and McKeown (2001) has been implemented to automatically estimate constraints on information ordering in the medical domain, at the content planning stage. Using a collection of semantically tagged transcripts written by domain experts, Duboue and McKeown (2001) identify basic adjacency patterns contained within a plan, as well as their ordering. MULTIGEN generates summaries of news on any topic. In such an unconstrained domain, it would be impossible to enumerate the semantics for all possible types of sentences which could match the elements of a schema, a plan or rhetorical relations. For instance, Duboue and McKeown build their content planner based on a set of 29 semantic categories; in our case, there is no such regularity in the input information. Furthermore, it would be difficult to specify a generic rhetorical plan for a summary of news. Instead, content determination in MULTIGEN is opportunistic, depending on the kinds of similarities that happen to exist between a set of news documents. Similarly, we describe here an ordering scheme that is opportunistic and bottom-up, depending on the cohesion and temporal connections that happen to exist between selected text.

Our ordering component takes place after the content selection of the information in a pipeline architecture, in contrast to generation systems, where usually the ordering and the content selection come in tandem. This separation might come at a cost — if there is no good ordering to the given extracted information, it is not possible to go back to the content selection to extract new information. In summarization, content selection is driven by salience criteria. We believe that the same ordering strategy should work with different content selectors, independently of their salience criteria. Therefore, we choose to keep the two components, selection and ordering, as two separate modules.

## 8. Conclusion and Future Work

In this paper we investigated information ordering constraints in multidocument summarization in the news genre. We evaluated two alternative ordering strategies, Chronological Ordering (CO) and Majority Ordering (MO). Our experiments show that MO performs well only when all input texts follow similar organization of the information. In the domains where this constraint holds, MO would be an appropriate and highly effective strategy. But in the news genre we cannot make this assumption; thus it is not an appropriate solution.

The Chronological Ordering (CO) algorithm can provide an acceptable solution for many cases, but is not sufficient when summaries contain information that is not event based. Our experiments, using a corpus that we collected of multiple alternative summaries each of multiple documents, show that cohesion is an important constraint contributing to ordering. Moreover, they also show that appropriate ordering of information is critical to allow for easy comprehension of the summary and that it is not the case that all possible orderings of information are acceptable. We developed an operational algorithm that integrates cohesion as part of the CO algorithm, and implemented it as part of the MULTIGEN summarization system. Our evaluation of the system shows significant improvement in summary quality.

While in this paper we focused on augmenting the CO algorithm, we believe that MO is a promising strategy and should not be neglected. It is clear that different forms of summarization are useful in different situations, depending on the intended purpose of the summary and on the types of documents summarized. For our future work, we plan to build on the approach we used for the DUC 2001 evaluation, where we developed a summarizer that would use different algorithms for summary generation depending on the type of input text. We suspect that ordering strategies may differ also, depending on the type of summary. Our work will first investigate whether we can use our augmented algorithm for other summary types. If the algorithm does not yield good orderings, we will investigate through corpus analysis other summary type specific constraints. We suspect that our augmented algorithm may apply, for instance, to biographical summaries, since the information being summarized is a mixture of event-based information that can be chronologically ordered along with descriptive information about the person. It is unclear whether it can apply to other types of summaries such as summaries of different events, since pieces of information may not be temporally related to each other. We also plan to identify the types of summaries which would benefit from using the MO algorithm or an augmented version of it (the same way the CO algorithm was augmented with the cohesion constraint).

## 9. Acknowledgments

This work was partially supported by DARPA grant N66001-00-1-8919, a Louis Morin scholarship and a Viros scholarship. We thank Eli Barzilay for providing help with the experiments interface, Michael Elhadad for the useful discussions and comments, and all the many voluntary participants in the experiments. Our initial work on the problem was presented at the Human Language Technologies Conference (San Diego, 2001). We also thank the anonymous reviewers of HLT and JAIR for their comments.